\documentclass[a4paper,fleqn]{cas-dc}

\usepackage[numbers]{natbib}

\usepackage{amsmath,amssymb,amsfonts,amsthm,mathtools,bm}
\usepackage{graphicx}
\usepackage{textcomp}
\usepackage{dsfont}
\usepackage{csquotes}
\usepackage{algorithm}
\usepackage{algpseudocode}
\usepackage{multirow,array}
\usepackage{csquotes}
\usepackage{booktabs}
\usepackage{xcolor}
\usepackage{url}
\usepackage{xurl}                                         %

\makeatletter
\newcommand\fs@betterruled{%
  \def\@fs@cfont{\bfseries}\let\@fs@capt\floatc@ruled
  \def\@fs@pre{\vspace*{5pt}\hrule height.8pt depth0pt \kern2pt}%
  \def\@fs@post{\kern2pt\hrule\relax}%
  \def\@fs@mid{\kern2pt\hrule\kern2pt}%
  \let\@fs@iftopcapt\iftrue}
\floatstyle{betterruled}
\restylefloat{algorithm}
\makeatother

\usepackage{circuitikz}
\usepackage{tikz}
\usetikzlibrary{arrows}
\usetikzlibrary{shapes}
\usetikzlibrary{shapes.geometric}
\usetikzlibrary{automata, arrows.meta, positioning}
\tikzset{
  invisible/.style={opacity=0},
  dots/.style={state,draw=none},
  visible on/.style={alt={#1{}{invisible}}},
   alt/.code args={<#1>#2#3}{%
     \alt<#1>{\pgfkeysalso{#2}}{\pgfkeysalso{#3}} %
  },
}

\newtheorem{lemma}{Lemma}

\newtheorem{theorem}{Theorem}
\newtheorem{corollary}{Corollary}
\newtheorem{assumption}{Assumption}

\newtheorem{proposition}{Proposition}
\theoremstyle{definition}   %
\newtheorem{example}{Example}
\newtheorem*{excont}{Example~\ref{ex:chain} (continued)}
\theoremstyle{plain}
\newtheorem{remark}{Remark}
\usepackage{textcomp}                                     %

\newcommand{\RR}{{\mathbb R}}

\newcommand{\M}{\mathcal M}
\newcommand{\X}{\mathcal X}
\newcommand{\U}{\mathcal U}
\newcommand{\F}{\mathcal F}
\newcommand{\G}{\mathcal G}

\newcommand{\Pp}{\mathbb P}
\newcommand{\E}{\mathbb E}
\newcommand{\1}{\mathds 1}

\newcommand{\Int}{\mathcal X^{E}}
\newcommand{\ind}[1]{\1\left\{#1\right\}}

\begin{document}
\let\WriteBookmarks\relax
\def\floatpagepagefraction{1}
\def\textpagefraction{.001}

\shorttitle{Rare Event Constrained Reinforcement Learning with Large Deviations}

\shortauthors{R. Misra}

\title [mode = title]{Rare Event Constrained Reinforcement Learning with Large Deviations}

\author[1]{Rahul Misra}

\cormark[1]

\ead{rmi@es.aau.dk}

\affiliation[1]{organization={Department of Electronic Systems, Automation and Control section, Aalborg University},
            addressline={Fredrik Bajers Vej 7 C},
            city={Aalborg East},
            postcode={9220},
            country={Denmark}}

\cortext[1]{Corresponding author}

\begin{abstract}
We study finite-horizon reinforcement learning with a safety constraint on the probability of reaching a failure set. In contrast to the well-studied safe reinforcement learning literature, we consider problems where the probability of reaching the failure set is vanishingly small, but the consequences of reaching the failure set are catastrophic. Direct simulation and estimation of the optimal policy is naturally ineffective when the failure probability is extremely small. Moreover, on a real system the controller can only choose actions, and this is not enough to make failures frequent. In contrast to the rare event simulation literature, our goal is not to estimate the probability of the rare event occurrence, but to upper bound it such that the failure constraint is satisfied. Thus, instead of sampling failures, we use the large deviations principle. We define a damage variable that on exceeding a certain threshold indicates failure. We study the cumulant generating function (CGF) as its positive root provides an upper bound on the failure probability via Cram\'er's theorem. We extend this approach to Markov decision processes with hitting times through state-action CGFs, and give finite-horizon and state-dependent versions of the bound. The state-action CGFs do not depend on the policy and can be estimated from regular transitions, so the bound can be checked for any policy a priori. This gives a policy improvement algorithm in which every policy satisfies the failure constraint with high probability.
\end{abstract}


\maketitle

\section{Introduction}
Safety-critical control problems are frequently posed as a cost minimization subject to a bound on the probability of a failure event. When the failure probability is of the order $10^{-6}$ or smaller (typically in aviation, nuclear, and medical systems), the failure constraint cannot be estimated via standard statistical techniques, as the number of required samples for observing a single failure is of the order of the reciprocal of the failure probability. Estimating a \emph{safe} optimal control policy via reinforcement learning is impossible, as it would require an infeasible amount of samples. Estimating probabilities of rare events, on the other hand, is relatively well studied \cite{asmussen2007,nair2022fundamentals,biondini2015}. The fundamental idea is to change the measure such that the rare event becomes likely, and then correct the bias. Intuitively, to study a rare disease it is more sensible to sample from a hospital with known occurrence(s) of the rare disease than from the general population. For a \emph{fixed} Markov chain, the zero-variance (optimal) change of measure is Doob's $h$-transform \cite{doob1957,asmussen2007}.

A classic example is Siegmund's algorithm \cite{siegmund1976}. To estimate the probability that a random walk $S_n=\sum_{i\leq n}\xi_i$ with negative drift ever climbs a level $c$, it samples the steps from the tilted law, $e^{\lambda^\star\xi}$ times the original transition law, where $\lambda^\star>0$ is the positive root of the cumulant generating function (CGF) $\kappa(\lambda)=\log\E\,[e^{\lambda\xi_1}]$ (Lundberg's coefficient). Under the tilted law, the walk drifts up and always reaches $c$, and the likelihood ratio $e^{-\lambda^\star S_\tau}$ is at most $e^{-\lambda^\star c}$, so the number of samples needed does not grow with $c$. All these methods \cite{ahamed2006,rubinstein2004,aguilar2024,rose2021,werner2026} need a simulator in which the transition law can be changed, and they consider a fixed chain, while in reinforcement learning the policy itself changes how rare the failure is. On a real system the controller can only choose actions, and this is not enough to make failures frequent in a useful way.

In contrast to the rare event simulation literature, our goal is not to estimate the probability of the rare event occurrence, but to upper bound it, as this is enough to satisfy the failure constraint. This is the idea behind our approach. We use Lundberg's inequality to bound the failure probability to be at most $e^{-\lambda^\star c}$ \cite{dembo1998,asmussen2010}. This bound needs only the CGF in steps, and can be estimated from normal operation. So instead of sampling from the tilted law, we estimate the Lundberg coefficient from data and use it to bound the constraint. This does not mean that failures play no role. We still need to know what a failure is, the damage potential, and the failure level (Assumption~\ref{as:phi}). What we do not need is to observe the failure event on the system we control. During normal operation the damage moves up and down, and from these steps we bound the probability of damage reaching the failure set. Without observed failures the failure probability cannot be estimated, but it can still be bounded from above. We extend this to Markov decision processes, where the transition law depends on the state action pair, through state-action CGFs and state-dependent Lundberg coefficients. The resulting bound is an exponential supermartingale, or a stochastic barrier certificate \cite{prajna2007,steinhardt2012}. Such certificates have been learned together with controllers \cite{mathiesen2023,zikelic2023}, usually with known dynamics and for regular probability values. Here the certificate comes from large deviations principle and it is estimated from data, unlike risk-sensitive control where the CGF is optimized as an exponential cost \cite{whittle1990,borkar2002}.

\smallskip
\noindent\textbf{Contributions.}
\begin{enumerate}
\item[(C1)] A safety certificate for MDPs with hitting times developed via state-action CGFs and state-dependent Lundberg coefficients (Theorem~\ref{thm:cert}, Corollaries~\ref{cor:cramer} and~\ref{cor:cells}).
\item[(C2)] We estimate of the safety certificate from trajectories (Lemma~\ref{lem:ucb}).
\item[(C3)] A simple learning algorithm that checks each new policy before it is run, so that every policy run during learning satisfies the constraint with high probability (Algorithm~\ref{alg:main}, Proposition~\ref{prop:safe}).
\end{enumerate}

\section{Preliminaries}\label{sec:prelim}
Consider a finite-horizon Markov decision process $\M = (\X, \U, C, P, T)$ defined on a finite state space $\X$, with finite control action space $\U$, having cost function $C:\X \times \U \to \RR_{\geq 0}$, and transition kernel $P:\X \times \U \to \Delta(\X)$. The control policy $\pi_\theta(u\mid x)$ is a distribution over the control action set $\U$, parametrized by parameter vector $\theta \in \RR^l$. We define a failure set $\F \subset \X$ and a goal set $\G \subset \X$. Both sets are absorbing and disjoint, $C=0$ on $\F\cup\G$, and we write $\Int:=\X\setminus(\F\cup\G)$ for the remaining states. For finite time $t$, the hitting time for $\F$ is
\begin{equation*}\label{eq:Hitting_time}
    \tau^t_\F:=\inf\{s\geq t:X_s\in\F\}, \quad \tau_\F := \tau^0_\F,
\end{equation*}
and the failure probability is $p_\theta(x_0):=\Pp_{\pi_\theta,x_0}(\tau_\F\leq T)$. Consider the following optimization, for a given $\delta \ll 1$,
\begin{equation}\label{eq:cmdp}
\begin{aligned}
    \min_{\theta}\ &J(\theta):= \E_{\pi_\theta}\left[\sum^{T-1}_{t=0}C(X_t,U_t)\mid X_0=x_0\right]\\
     &\text{s.t.}\quad p_\theta(x_0)\leq\delta.
\end{aligned}
\end{equation}
\noindent\textbf{Setting.} The kernel $P$ is unknown. The controller runs a policy on the system and observes the states, actions and costs. It cannot start the system from a state of its choice and it cannot change the randomness of $P$. The constraint in \eqref{eq:cmdp} needs to be satisfied by all control policies at all times. Crude Monte Carlo requires $\mathcal{O}(1/p_\theta)$ trajectories merely to observe failures and $\mathcal{O}(1/(\epsilon^2 p_\theta))$ trajectories for estimating $p_\theta$ up to relative error $\epsilon$ \cite{asmussen2007}. Failure is usually the end of an accumulation, such as wear, energy or the distance to a limit. We describe this as the following damage potential $\phi$.
\begin{assumption}\label{as:phi}
There exist $\phi:\X\to\RR$ and $b\in\RR$ with $\F=\{x:\phi(x)\geq b\}$, and the increments $\xi_{t+1}:=\phi(X_{t+1})-\phi(X_t)$ satisfy $|\xi_{t+1}|\leq\xi_{\max}$ for a known constant $\xi_{\max}$.
\end{assumption}
We define $c:=b-\phi(x_0)$ as the distance to failure. The constant $\xi_{\max}$ is the maximum possible damage in one step. This is usually a physical limit, such as the maximum wear or energy change per time step. Assumption \ref{as:phi} is required for the following reasons. Firstly, it makes the CGFs of Section~\ref{sec:ld} finite, so that the Lundberg coefficient exists (light tails). Secondly, it gives the known range of $e^{\lambda\xi}$ that is needed to learn the CGF from data in Lemma~\ref{lem:ucb}.

We shall now define the failure $h$-function. For a fixed policy $\pi_{\theta}$, we define the finite-horizon failure $h$-function,
\begin{equation*}\label{eq:h_function}
    h^\theta_{t}(x) = \Pp_{\pi_{\theta}}\left(\tau^t_{\F}\leq T\mid X_t = x \right).
\end{equation*}
The $h$-function gives the probability of eventual failure before the specified horizon, from the current state and time. For any running time $t$, where $0 \leq t \leq T$,
\begin{equation}\label{eq:failure-boundary}
    h^\theta_{t}(x) = 1, \text{ if } x \in \F, \quad h^\theta_{t}(x) = 0, \text{ if } x \in \G.
\end{equation}
At the terminal time $T$, for a nonfailure state $x\notin\F$,
\begin{equation}\label{eq:terminal-boundary}
    h^\theta_{T}(x)=0.
\end{equation}
For $x\in\Int$ and $t<T$, the $h$-function satisfies the following backward recursion,
\begin{equation}\label{eq:h_function-recursion}
\begin{aligned}
    h^{\theta}_t(x)&= \sum_{x'\in\X} P_\theta(x'\mid x)h^{\theta}_{t+1}(x')\\
    &= \sum_{u\in\U}\pi_\theta(u\mid x) g^{\theta}_t(x,u),
\end{aligned}
\end{equation}
where $P_\theta(x'\mid x):=\sum_{u\in\U}\pi_\theta(u\mid x)P(x'\mid x,u)$ is the kernel and $g^{\theta}_t(x,u)$ is the state-action $h$-function
\begin{equation*}\label{eq:g_function}
    g^{\theta}_t(x,u) := \sum_{x'\in\X} P(x'\mid x,u)\, h^\theta_{t+1}(x').
\end{equation*}
Thus, $h^\theta$ is harmonic (or satisfies the recursion \eqref{eq:h_function-recursion}), and the failure probability appearing in the constraint is $p_{\theta}(x_0)=h^\theta_{0}(x_0)$. For a fixed $\theta$, the optimal zero-variance change of measure for estimating $h^\theta_0(x_0)$ is the kernel,
\begin{equation}\label{eq:doob}
    S^{\theta,\star}_t(x'\mid x) := P_\theta(x'\mid x) \frac{h^{\theta}_{t+1}(x')}{h^{\theta}_t(x)}, \quad h^{\theta}_t(x)>0.
\end{equation}
As $S^{\theta,\star}$ is conditioned on the failure event happening, every trajectory hits $\F$. Note that the controller can only choose $\theta$ and cannot obtain $S^{\theta,\star}$ as it depends on the underlying environment. The kernel \eqref{eq:doob} can be split into a part the controller can control and a part it cannot control. For $h^\theta_t(x)>0$, define the tilted policy
\begin{equation*}\label{eq:tilt_pi}
    \tilde\pi^{\theta}_t(u\mid x) := \pi_\theta(u\mid x)\frac{g^{\theta}_t(x,u)}{h^{\theta}_t(x)},
\end{equation*}
which is a distribution by \eqref{eq:h_function-recursion}. Multiplying it by $P(x'\mid x,u)h^\theta_{t+1}(x')/g^\theta_t(x,u)$ and summing over $u$ gives \eqref{eq:doob}. So the Doob transform first tilts the action and then tilts the environment. In practice, the controller can sample its actions from $\tilde\pi^\theta_t$, but it cannot tilt the environment.

\begin{example}[Damage chain]\label{ex:chain}
We use the following example throughout the paper to illustrate the theory. To illustrate the policy dependence of the failure $h$-function, consider $\mathcal{M}$ with $\X=\{0,1,\dots,N\}$. The state $X_t$ represents accumulated damage for an expensive machine such as a wind turbine. The target set is $\G=\{0\}$, corresponding to safe termination, and the failure set is $\F=\{N\}$, corresponding to catastrophic failure. Both are absorbing. At each interior state $x\in\{1,\dots,N-1\}$ the controller selects an action from $\U=\{\mathsf{s},\mathsf{r}\}$, where $\mathsf{s}$ denotes a safe operating mode and $\mathsf{r}$ a less costly but riskier mode, with $C(x,\mathsf{r}) =: C_{\mathsf r} < C_{\mathsf s} =: C(x,\mathsf{s})$. The policy is
\begin{equation*}\label{eq:expolicy}
    \pi_\theta(\mathsf{r}\mid x)=\theta,\quad \pi_\theta(\mathsf{s}\mid x)=1-\theta,\quad \theta\in[0,1].
\end{equation*}
The expected one-step cost is $\bar C(\theta) = \theta C_{\mathsf r} + (1-\theta)C_{\mathsf s}$. The induced probability of transition towards failure is
\begin{equation*}\label{eq:qtheta}
    q_\theta = (1-\theta)q_{\mathsf{s}} + \theta q_{\mathsf{r}}, \quad q_{\mathsf s} < q_{\mathsf r} < \tfrac12,
\end{equation*}
Therefore, for an interior state,
\begin{equation*}\label{eq:exkernel}
    P_\theta(x'\mid x) =
\begin{cases}
    q_\theta, & x'=x+1,\\
    1-q_\theta, & x'=x-1,\\
    0,&\text{otherwise.}
\end{cases}
\end{equation*}
From \eqref{eq:failure-boundary}, \eqref{eq:terminal-boundary}, \eqref{eq:h_function-recursion}, the $h$-function satisfies the conditions,
\begin{equation*}\label{eq:exbc}
    h^{\theta}_t(N)=1,\quad h^{\theta}_t(0)=0,\quad h^{\theta}_T(x)=0\ \ \forall x\neq N,
\end{equation*}
and for $x\in\{1,\dots,N-1\}$, $t<T$,
\begin{equation*}\label{eq:exrec}
    h^{\theta}_t(x) = q_\theta h^{\theta}_{t+1}(x+1) + (1-q_\theta)h^{\theta}_{t+1}(x-1).
\end{equation*}
This MDP satisfies Assumption~\ref{as:phi} with $\phi(x)=x$, $b=N$ and $\xi_{\max}=1$, and the distance to failure is $c=N-x_0$. Increasing $\theta$ lowers the cost per step but makes episodes longer. The risky mode is optimal only if $C_{\mathsf r}/C_{\mathsf s}<(1-2q_{\mathsf r})/(1-2q_{\mathsf s})$. We take $N=20$, $x_0=2$, $T=100$, $q_{\mathsf s}=0.2$, $q_{\mathsf r}=0.4$, $C_{\mathsf s}=1$ and $C_{\mathsf r}=0.25$, which satisfies this condition. Then $p_\theta(x_0)$ grows from $1.36\times10^{-11}$ at $\theta=0$ to $3.08\times10^{-4}$ at $\theta=1$, and $J$ falls from $3.333$ to $2.457$. With $\delta=10^{-5}$ the exact optimum is $\theta^\star=0.7505$ with $J=2.914$. Crude Monte Carlo needs about $10^{7}$ episodes to estimate $p_\theta(x_0)$ to $10\%$ near $\theta^\star$. Both modes move the damage up with positive probability, so every action can lead to failure, and sampling the actions from $\tilde\pi^\theta_t$ does not reduce the variance.
\end{example}

\section{Large Deviations Principle}\label{sec:ld}
This section summarizes useful large deviation concepts. We first state them for independent steps and then define their state-dependent versions for the MDP. Let $\xi_1,\xi_2,\dots$ be i.i.d., $S_n=\sum_{i\leq n}\xi_i$, and let the cumulant generating function (CGF) be
\begin{equation*}\label{eq:cgf}
    \kappa(\lambda):=\log\E\,[e^{\lambda\xi_1}].
\end{equation*}
For simplicity we assume $\kappa(\lambda)<\infty$ for all $\lambda\in\RR$, which holds under Assumption~\ref{as:phi}. $\kappa$ is convex, $\kappa(0)=0$ and $\kappa'(0)=\E[\xi_1]$. Its Legendre transform,
\begin{equation}\label{eq:rate}
    I(v):=\sup_{\lambda\in\RR}\{\lambda v-\kappa(\lambda)\}
\end{equation}
is the rate function. It is convex, nonnegative, and zero only at $v=\E[\xi_1]$. The next theorem quantifies the probability of observing large deviations from the expected value.
\begin{theorem}[Cram\'er {\cite[Thm.~3.12]{nair2022fundamentals}, \cite[Thm.~2.2.3]{dembo1998}}]\label{thm:cramer}
For $v>\E[\xi_1]$,
\begin{equation*}
    \lim_{n\to\infty}-\tfrac1n\log\Pp(S_n\geq nv)=I(v).
\end{equation*}
\end{theorem}
The proof of Theorem~\ref{thm:cramer} is in \cite[Sec.~3.4]{nair2022fundamentals} and \cite[Sec.~2.2]{dembo1998}. The idea is a change of measure that makes the rare event likely \cite[Sec.~VI.2]{asmussen2007}. Let $F$ be the law of $\xi_1$ and, for $\lambda\geq0$, define the exponentially tilted law $F_\lambda(d\xi):=e^{\lambda\xi-\kappa(\lambda)}F(d\xi)$. It is a probability law because $\E[e^{\lambda\xi_1}]=e^{\kappa(\lambda)}$, and its mean is $\kappa'(\lambda)$. Since $\kappa$ is convex, $\kappa'$ is increasing, so a larger $\lambda$ puts more weight on large steps. Choose $\lambda_v\geq0$ with $\kappa'(\lambda_v)=v$. Under $F_{\lambda_v}$ the mean step is $v$, so $\{S_n\geq nv\}$ is no longer rare and by the central limit theorem its probability tends to $1/2$. The price of the change is the likelihood ratio of a path,
\begin{equation*}
    \prod_{i=1}^{n}\frac{F(d\xi_i)}{F_{\lambda_v}(d\xi_i)}=e^{-\lambda_vS_n+n\kappa(\lambda_v)},
\end{equation*}
which equals $e^{-n(\lambda_vv-\kappa(\lambda_v))}=e^{-nI(v)}$ on paths with $S_n=nv$, since the supremum in \eqref{eq:rate} is attained at $\lambda_v$. So $\Pp(S_n\geq nv)$ is about $e^{-nI(v)}$ times a probability that is not small. In other words, the rare event happens in the \emph{least unlikely} way, as all steps behave as if their mean were $v$, and $I(v)$ is the cost (or energy required) per step of this change. It is also the relative entropy of $F_{\lambda_v}$ with respect to $F$. In our problem setting, we are interested in the damage which might increase by $c$ at some time $n$, instead of the average damage. Therefore, we need to modify the rare event $\{S_n\geq nv\}$ to $\{\exists n:  S_n\geq c\}$, and this is done via Lundberg's coefficient. Assume $\E[\xi_1]<0$ and $\Pp(\xi_1>0)>0$. Then $\kappa$ has a second root $\lambda^\star>0$,
\begin{equation}\label{eq:lundberg}
    \kappa(\lambda^\star)=0,
\end{equation}
called the Lundberg coefficient \cite{asmussen2010}, also known as the Cram\'er root or the adjustment coefficient \cite[Sec.~VI.2a]{asmussen2007}. Since $\E[e^{\lambda^\star\xi_1}]=1$, the process $e^{\lambda^\star S_n}$ is a martingale, and stopping it when $S_n \geq c$ gives Lundberg's inequality
\begin{equation}\label{eq:lundbergineq}
    \Pp(\exists n:S_n\geq c)\leq e^{-\lambda^\star c}.
\end{equation}
The next lemma connects the Lundberg coefficient to the rate function $I(v)$ in \eqref{eq:rate}.
\begin{lemma}\label{lem:speed}
$\lambda^\star=\min_{v>0}I(v)/v$, and the minimum is attained at $v^\star=\kappa'(\lambda^\star)$.
\end{lemma}

\begin{proof}
By \eqref{eq:rate} with $\lambda=\lambda^\star$, $I(v)\geq\lambda^\star v-\kappa(\lambda^\star)=\lambda^\star v$ for all $v$. At $v^\star=\kappa'(\lambda^\star)$ the supremum in \eqref{eq:rate} is attained at $\lambda^\star$, so $I(v^\star)=\lambda^\star v^\star$.
\end{proof}

Lemma~\ref{lem:speed} explains \eqref{eq:lundbergineq} with Cram\'er's theorem. A path that increases damage to $c$ at speed $v$ needs $c/v$ steps and has probability about $e^{-(c/v)I(v)}$. The cheapest speed is $v^\star$, and its cost is $\lambda^\star c$. The tilted transition law, $e^{\lambda^\star\xi}$ times the original law, has mean $v^\star$. This is the idea behind Siegmund's algorithm \cite{siegmund1976}.

\begin{excont}
Under $\pi_\theta$ the damage moves by $\pm1$ with up-probability $q=q_\theta$, so $\kappa(\lambda)=\log\big(qe^{\lambda}+(1-q)e^{-\lambda}\big)$ and $\lambda^\star=\log\big((1-q)/q\big)$. Tilting a step by $e^{\lambda^\star\xi}$ results in up-probability $1-q$ and mean $v^\star=1-2q$ and this is the harmonic Doob $h$-transform with $h(x)=e^{\lambda^\star(x-b)}$ \eqref{eq:lundberg}.
\end{excont}

\begin{figure*}[t]
\centering
\includegraphics[width=0.75\textwidth]{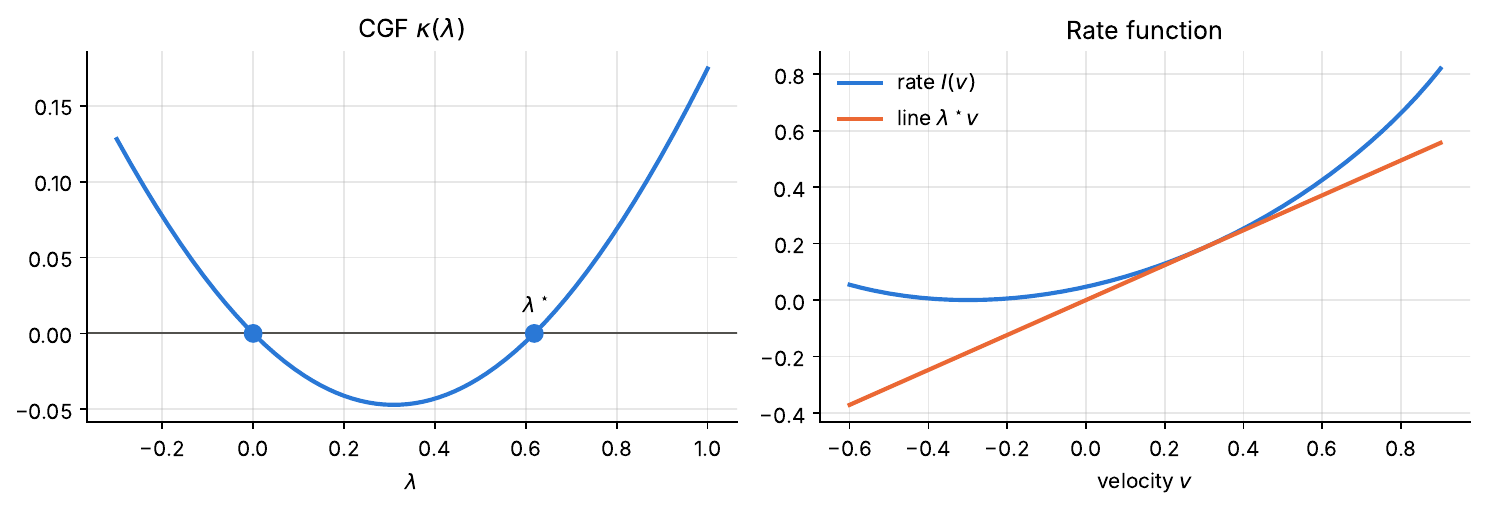}
\caption{Left: CGF of one step of Example~\ref{ex:chain} with $q=0.35$ and its positive root, the Lundberg coefficient $\lambda^\star=0.619$. Right: the rate function $I$ and the line $\lambda^\star v$, which touches it at the cheapest speed $v^\star=1-2q=0.30$ (Lemma~\ref{lem:speed}).}
\label{fig:cgf}
\end{figure*}
We now drop the i.i.d assumption and consider controlled Markov chains or MDP's. Now the CGF depends on the state and the action. For $x\in\Int$, $u\in\U$ and $\lambda\in\RR$ we define the state-action CGF and the policy CGF
\begin{align}
    \kappa(x,u;\lambda)&:=\log\E\big[e^{\lambda\xi_{t+1}}\,\big|\,X_t=x,U_t=u\big],\notag\\
    \label{eq:polcgf}
    \kappa_\theta(x;\lambda)&:=\log\sum_u\pi_\theta(u\mid x)\,e^{\kappa(x,u;\lambda)}.
\end{align}
 $\kappa_\theta(x;\cdot)$ is the CGF of the next $\xi_{t+1}$ under $\pi_\theta$ and is convex in $\lambda$. The state-dependent Lundberg coefficient is
\begin{equation*}
    \lambda^\star_\theta(x):=\sup\{\lambda\geq0:\kappa_\theta(x;\lambda)\leq0\}.
\end{equation*}
It is positive when the damage drifts down at $x$, and zero when it drifts up. By convexity, $\kappa_\theta(x;\lambda)\leq0$ for all $\lambda\in[0,\lambda^\star_\theta(x)]$. The central idea is to replace the single CGF in i.i.d case with a state dependent CGF that is only conditioned on the current state due to the Markov property.

\section{The safety certificate}\label{sec:cert}
The next theorem specifies the conditions for constructing a safety certificate $W$ that bounds $p_\theta(x_0)$.
\begin{theorem}\label{thm:cert}
Let $W_t:\X \to [-\infty,\infty)$, $t=0,\dots,T$, and $w_{\F}\in\RR$ satisfy
\begin{align}
    &\sum_u\pi_\theta(u\mid x)\sum_{x'}P(x'\mid x,u)\,e^{W_{t+1}(x')-W_t(x)}\leq1, \label{eq:super} \\
    &\begin{aligned}
        &W_t(x)\geq w_{\F}\quad (x\in\F),\\
        &W_{t+1}(x)\leq W_t(x)\quad (x\in\G),
    \end{aligned} \label{eq:bdry}
\end{align}
where \eqref{eq:super} is required for $x\in\Int$ and $t<T$. Then
\begin{equation*}
    p_\theta(x_0)\leq\exp\big(W_0(x_0)-w_{\F}\big) \quad \text{for every } T.
\end{equation*}
Equality holds for $W_t=\log h^\theta_t$ and $w_\F=0$, with \eqref{eq:super} multiplied by $e^{W_t(x)}$.
\end{theorem}

\begin{proof}
Let $\mathcal H_t:=(X_0,U_0,\dots,U_{t-1},X_t)$ be the history up to time $t$ and consider the process stopped at failure,
\begin{equation*}
    Z_t:=\exp \left(W_{t\wedge\tau_\F}\left(X_{t\wedge\tau_\F}\right)\right),\quad t=0,\dots,T,
\end{equation*}
where $t\wedge\tau_\F$ is the minimum between $t$ and $\tau_\F$. It is nonnegative and $Z_0=e^{W_0(x_0)}$. We first show that $\E[Z_{t+1}\mid\mathcal H_t]\leq Z_t$ for $t<T$. If $\tau_\F\leq t$, the process is stopped and $Z_{t+1}=Z_t$. If $X_t=x\in\Int$, then $\tau_\F>t$ and $Z_{t+1}=e^{W_{t+1}(X_{t+1})}$, also when $X_{t+1}\in\F$, since then $\tau_\F=t+1$. Given $\mathcal H_t$, the action is drawn from $\pi_\theta(\cdot\mid x)$ and the next state from $P(\cdot\mid x,u)$, so by \eqref{eq:super},
\begin{equation*}
    \begin{aligned}
    \E[Z_{t+1}\mid\mathcal H_t]&=\sum_u\pi_\theta(u\mid x)\sum_{x'}P(x'\mid x,u)\,e^{W_{t+1}(x')}\\
    &\leq e^{W_t(x)}=Z_t.
    \end{aligned}
\end{equation*}
If $X_t=x\in\G$, then $X_{t+1}=x$ because $\G$ is absorbing, and $Z_{t+1}=e^{W_{t+1}(x)}\leq e^{W_t(x)}=Z_t$ by the second condition in \eqref{eq:bdry}. Since $X_t\in\F$ implies $\tau_\F\leq t$, these three cases cover all paths, so $Z_t$ is a supermartingale. Taking expectations and repeating over $t$ gives $\E[Z_T]\leq\E[Z_{T-1}]\leq\dots\leq Z_0=e^{W_0(x_0)}$. On the failure event $\{\tau_\F\leq T\}$ we have $Z_T=e^{W_{\tau_\F}(X_{\tau_\F})}\geq e^{w_\F}$ by the first condition in \eqref{eq:bdry}. Hence
\begin{equation*}
    \begin{aligned}
    e^{w_\F}p_\theta(x_0)&=e^{w_\F}\Pp(\tau_\F\leq T)\leq\E\big[Z_T\ind{\tau_\F\leq T}\big]\\
    &\leq\E[Z_T]\leq e^{W_0(x_0)},
    \end{aligned}
\end{equation*}
which is the claim. No limit in $T$ is taken, so the bound holds for every finite horizon. For $W_t=\log h^\theta_t$ and $w_\F=0$, \eqref{eq:failure-boundary} gives $W_t=0$ on $\F$ and $W_t=-\infty$ on $\G$, so \eqref{eq:bdry} holds. Multiplied by $e^{W_t(x)}=h^\theta_t(x)$, \eqref{eq:super} is the recursion \eqref{eq:h_function-recursion}, which holds with equality. The bound is then $h^\theta_0(x_0)=p_\theta(x_0)$.
\end{proof}
Condition \eqref{eq:super} says that $\E_\theta[e^{W_{t+1}(X_{t+1})}\mid X_t = x] \leq e^{W_t(x)}$, or $e^{W_t(x)}$ is a supermartingale, and we can upper bound the failure probability. Now, for constructing a certificate, we have to choose $W$ in terms of damage $\phi$ (Assumption \ref{as:phi}), and here choosing $W(x_t) = \lambda(\phi(x_t)-b) = -\lambda c$ satisfies \eqref{eq:polcgf}, and substituting it in condition \eqref{eq:super} results in
\begin{equation}\label{eq:State_lundberg}
    \E_\theta[e^{\lambda \xi_{t+1}}\mid X_t = x] \leq 1, \text{ or } \kappa_\theta(x;\lambda) \leq 0.
\end{equation}
For an MDP, the increments $\xi_{t+1}$ depends on the state and there may not be a single $\lambda > 0$ satisfying \eqref{eq:State_lundberg}. The following two corollaries address this.
\begin{corollary}\label{cor:cramer}
Let $\bar\kappa_\theta(\lambda):=\max_{x\in\Int}\kappa_\theta(x;\lambda)$. Under Assumption~\ref{as:phi},
\begin{equation}\label{eq:cramerbound}
    p_\theta(x_0)\leq B_\theta:=\inf_{\lambda\geq0}\exp\Big(-\lambda c+T\,
    \big[\bar\kappa_\theta(\lambda)\big]^+\Big).
\end{equation}
In particular, if $\bar\kappa_\theta(\lambda^\star)=0$ for some $\lambda^\star>0$, then $p_\theta(x_0)\leq e^{-\lambda^\star c}$ for every $T$.
\end{corollary}

\begin{proof}
Fix $\lambda\geq0$, write $\bar\kappa^+:=[\bar\kappa_\theta(\lambda)]^+$, let $w_\F=0$, and
\begin{equation*}
    W_t(x)=\lambda\big(\phi(x)-b\big)+(T-t)\,\bar\kappa^+,\quad x\in\X.
\end{equation*}
We now check the conditions of Theorem~\ref{thm:cert}. On $\F$ we have $\phi(x)\geq b$, so $W_t(x)\geq0=w_\F$. For every $x$, $W_{t+1}(x)-W_t(x)=-\bar\kappa^+\leq0$, thus satisfying \eqref{eq:bdry}. For $x\in\Int$ and $t<T$, $W_{t+1}(x')-W_t(x)=\lambda\xi-\bar\kappa^+$ with $\xi=\phi(x')-\phi(x)$, so by the definition \eqref{eq:polcgf} the left side of \eqref{eq:super} is
\begin{equation*}
    \begin{aligned}
    &e^{-\bar\kappa^+}\sum_u\pi_\theta(u\mid x)\,\E\big[e^{\lambda\xi_{t+1}}\,\big|\,X_t=x,U_t=u\big]\\
    &=e^{\kappa_\theta(x;\lambda)-\bar\kappa^+}\leq 1, \quad \text{since }\kappa_\theta(x;\lambda)\leq\bar\kappa_\theta(\lambda)\leq\bar\kappa^+.
    \end{aligned}
\end{equation*}
As $\phi(x_0)-b=-c$, Theorem~\ref{thm:cert} gives $p_\theta(x_0)\leq e^{W_0(x_0)}=\exp\big(-\lambda c+T\bar\kappa^+\big)$. This holds for every $\lambda\geq0$, which gives \eqref{eq:cramerbound}. If for $\lambda^\star > 0$, $\bar\kappa_\theta(\lambda^\star)=0$, then $\bar\kappa^+=0$, and the bound becomes $e^{-\lambda^\star c}$, which no longer depends on $T$.
\end{proof}
Thus, if there exists a positive $\lambda^\star$, we can obtain a time-independent bound. Otherwise, a larger $\lambda$ produces a lower bound for small $T$. When $\kappa_\theta(x;\cdot)$ does not depend on $x$ and $\G$ lies below $x_0$, the damage is a random walk until absorption, and \eqref{eq:cramerbound} is tight. If $c$ and $T$ grow with $T/c$ fixed, $-\frac1c\log p_\theta(x_0)$ and $-\frac1c\log B_\theta$ have the same limit. This is the finite-time ruin asymptotics of \cite[Ch.~V]{asmussen2010}, and Section~\ref{sec:num} checks it numerically. The maximum over states makes \eqref{eq:cramerbound} conservative when the dynamics change with the damage. Splitting the states into cells with common state-dependent Lundberg coefficients removes most of the conservatism. Another reason for introducing cells is that it allows us to group states based on damage $\phi(x)$, and in normal operation the failure states might not be visited often, so we can combine safe states with similar CGF in a cell. The splitting of states into cells will be based on a decay rate $s$, and this is stated in the following corollary.

\begin{corollary}\label{cor:cells}
Let $s:\RR\to[0,\infty)$ be nonincreasing with $s(\phi(x))\leq\lambda^\star_\theta(x)$ for all $x\in\Int$. Then for every $T$,
\begin{equation}\label{eq:decaybound}
    p_\theta(x_0)\leq\exp\Big(-\int_{\phi(x_0)}^{b}s(y)\,dy\Big).
\end{equation}
\end{corollary}

\begin{proof}
Let $\psi(y):=\int_b^y s(z)\,dz$, and take $W_t(x)=\psi(\phi(x))$ for all $t$ and $w_\F=0$. On $\F$, $\phi(x)\geq b$ and $s\geq0$ give $W_t(x)\geq0$, and the condition on $\G$ holds with equality since $W_t$ does not depend on $t$. Let $x\in\Int$ and $y=\phi(x)$. Since $s$ is nonincreasing, $s(z)\leq s(y)$ for $z>y$ and $s(z)\geq s(y)$ for $z<y$, so for every $y'$,
\begin{equation*}
    \psi(y')-\psi(y)=\int_y^{y'}s(z)\,dz\leq s(y)\,(y'-y).
\end{equation*}
With $y'=\phi(x')$ it gives $W_{t+1}(x')-W_t(x)\leq s(\phi(x))\,\xi$, $\xi=\phi(x')-\phi(x)$, so the left side of \eqref{eq:super} is at most $e^{\kappa_\theta(x;s(\phi(x)))}$. Since $0\leq s(\phi(x))\leq\lambda^\star_\theta(x)$, $\kappa_\theta(x;s(\phi(x)))\leq0$ and \eqref{eq:super} holds. Theorem~\ref{thm:cert} gives $p_\theta(x_0)\leq e^{\psi(\phi(x_0))}$, which is \eqref{eq:decaybound}.
\end{proof}
By \eqref{eq:decaybound}, each unit of distance from the failure level at damage $y$ lowers the bound by about the factor $e^{-s(y)}$. In practice, we use cells. Let $-\infty=b_0<b_1<\dots<b_K=b$ and define cell $D_k:=\{x\in\Int:b_{k-1}\leq\phi(x)<b_k\}$. Let $s=s_k$ on $[b_{k-1},b_k)$, $s=s_K$ on $[b,\infty)$, with $s_1\geq s_2\geq\dots\geq s_K\geq0$, and $s_k\leq\lambda^\star_\theta(x)$ for all $x\in D_k$. Then \eqref{eq:decaybound} is,
\begin{equation}\label{eq:cellbound}
    p_\theta(x_0)\leq\exp\Big(-\sum_{k=1}^K s_k\,\ell_k\Big),
\end{equation}
where $\ell_k$ is the length of $[b_{k-1},b_k]\cap[\phi(x_0),b]$. A cell $D_k$ where nothing is known can be given $s_k=0$, because $\kappa_\theta(x;0)=0$ for any dynamics. The bound then loses the part that this cell would add, but it stays valid. Since the decay rates may not increase, all cells above it also get $s=0$.

\begin{excont}
Here $\kappa(x,u;\lambda)=\log\big(q_ue^{\lambda}+(1-q_u)e^{-\lambda}\big)$ does not depend on $x$, and the policy CGF is $\kappa_\theta(x;\lambda)=\log\big(q_\theta e^{\lambda}+(1-q_\theta)e^{-\lambda}\big)$ for every $x\in\Int$. So the maximum over $x$ in Corollary~\ref{cor:cramer} is not conservative. Table~\ref{tab:cert} compares the bound \eqref{eq:cramerbound} with the exact failure probability.
\end{excont}

\section{Estimating the certificate from data}\label{sec:data}
The certificate needs only the state-action CGFs, and these do not depend on the policy. So transitions observed under any policy can be used to check any other policy. We split $\Int$ into cells $D_1,\dots,D_K$ and assume the following.

\begin{assumption}\label{as:cells}
For each cell $D_k$ and action $u$, the distribution of next increment $\xi_{t+1}$ given $X_t=x$, $U_t=u$ is the same for all $x\in D_k$ and all $t$.
\end{assumption}

Assumption~\ref{as:cells} allows us to define $\kappa(D_k,u;\lambda)$ as the common state-action CGF on $D_k$ or the cell CGF.

\begin{excont}
The up-probability of each mode does not depend on the state, so one cell $D_1=\Int$ satisfies Assumption~\ref{as:cells}. Every step taken with mode $u$ is then a sample of $\kappa(D_1,u;\cdot)$, whichever policy chose the mode.
\end{excont}
A cell can be a single state, and then Assumption~\ref{as:cells} is the Markov property. Under Assumption~\ref{as:cells} and the Markov property, the increments $Y_1,Y_2,\dots$ seen right after successive uses of action $u$ in cell $D_k$ (not the increments $\xi$ along the path) are i.i.d., for any policy that chose these actions (due to Doob's optional skipping theorem \cite{doob1936}). Let $\hat M_{k,u,n}(\lambda)$ be the sample mean $(1/n)\sum^n_{j=1}e^{\lambda Y_j}$, and fix a grid $\Lambda_{\mathrm g}\subset[0,\lambda_{\max}]$ with $L$ points.

\begin{lemma}\label{lem:ucb}
Under Assumptions~\ref{as:phi} and~\ref{as:cells}, with probability at least $1-\eta$, for all $k$, $u$, $\lambda\in\Lambda_{\mathrm g}$ and $n\geq1$ at once,
\begin{equation}\label{eq:ucb}
\begin{aligned}
    e^{\kappa(D_k,u;\lambda)}&\leq\hat M_{k,u,n}(\lambda)\\
    &\quad+\big(e^{\lambda\xi_{\max}}-e^{-\lambda\xi_{\max}}\big)
    \sqrt{\tfrac{\log(|\U|KLn(n+1)/\eta)}{2n}}.
\end{aligned}
\end{equation}
\end{lemma}

\begin{proof}
Fix $k$ and $u$, and let $Y_j$ be the increment observed the $j$-th time that action $u$ is used in cell $D_k$. Given everything observed before that time, $Y_j$ has the law of Assumption~\ref{as:cells}, by the Markov property. So $Y_j$ is independent of $Y_1,\dots,Y_{j-1}$ and has the same law, whatever policy chose the actions, also when the policy was chosen from earlier data. Hence $Y_1,Y_2,\dots$ are i.i.d.\ with $\E[e^{\lambda Y_j}]=e^{\kappa(D_k,u;\lambda)}$, and $\hat M_{k,u,n}(\lambda)$ is the mean of the first $n$ values of $e^{\lambda Y_j}$. By Assumption~\ref{as:phi}, $e^{\lambda Y_j}\in[e^{-\lambda\xi_{\max}},e^{\lambda\xi_{\max}}]$, an interval of length $\Delta_\lambda:=e^{\lambda\xi_{\max}}-e^{-\lambda\xi_{\max}}$. Hoeffding's inequality \cite[Thm.~2]{hoeffding1963}, applied to $-e^{\lambda Y_j}$, gives, for each $n$ and $\eta'\in(0,1)$,
\begin{equation*}
    \Pp\Big(e^{\kappa(D_k,u;\lambda)}>\hat M_{k,u,n}(\lambda)+\Delta_\lambda\sqrt{\tfrac{\log(1/\eta')}{2n}}\Big)\leq\eta'.
\end{equation*}
We take $\eta'=\eta/(|\U|KLn(n+1))$. A union bound over the $|\U|KL$ choices of $(k,u,\lambda)$ and over $n\geq1$ bounds the probability that \eqref{eq:ucb} fails anywhere by $\eta\sum_{n\geq1}1/(n(n+1))=\eta$.
\end{proof}

Write $\kappa^{\mathrm u}(D_k,u;\lambda)$ for the logarithm of the right side of \eqref{eq:ucb}, with $\kappa^{\mathrm u}=+\infty$ if action $u$ was never used in $D_k$. Since \eqref{eq:polcgf} increases with each $e^{\kappa(x,u;\lambda)}$, using $\kappa^{\mathrm u}$ in place of $\kappa$ gives a policy CGF $\kappa^{\mathrm u}_\theta$ that is an upper bound of $\kappa_\theta$. We define the data certificate $B^{\mathrm u}_\theta$ as \eqref{eq:cramerbound} with $\kappa^{\mathrm u}_\theta$ and the minimum over $\Lambda_{\mathrm g}$, or as \eqref{eq:cellbound} with the decay rates $s_k=\min_{i\leq k}\max\{\lambda\in\Lambda_{\mathrm g}:\kappa^{\mathrm u}_\theta(D_i;\lambda)\leq0\}$. On the event of Lemma~\ref{lem:ucb}, $p_\theta(x_0)\leq B^{\mathrm u}_\theta$ \emph{for all $\theta$ at the same time}, whatever data were used. It is important to note the following limitations. A policy that uses an action never tried in some cell cannot be certified. A cell that was never visited gets decay rate zero, so the certificate becomes larger but not optimistic. Because \eqref{eq:ucb} holds for all $n$, data can be added while the policy changes. However, an increment larger than $\xi_{\max}$ that was never seen would invalidate the bound (Assumption~\ref{as:phi}). Thus, prior knowledge of $\xi_{\max}$ is needed by any estimate of a probability that lies far below what the data show.

\begin{figure}
\centering
\includegraphics[width=\columnwidth]{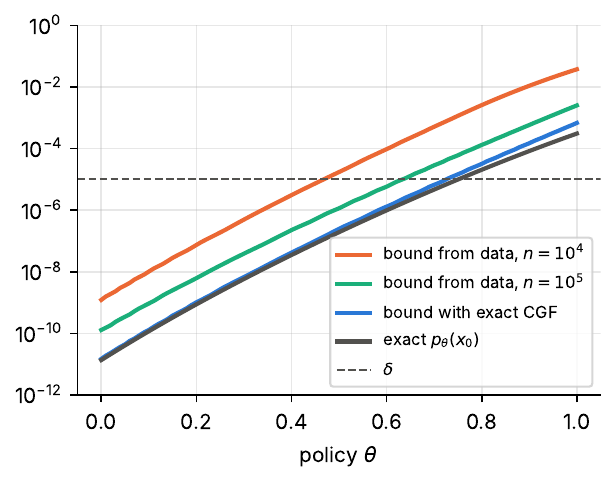}
\caption{Bound on $p_\theta(x_0)$ for every policy $\theta$ in Example~\ref{ex:chain} ($T=100$), from one data set of $n$ steps of the policy $\theta=0.5$ and with the exact CGFs, together with the exact failure probability. The largest certified policy is where the bound crosses $\delta$.}
\label{fig:bound}
\end{figure}

\section{Certified Policy Improvement}\label{sec:opt}
The certificate can be computed for any $\theta$ from the data, before running $\pi_\theta$, and can be used for learning without running unsafe policies, if the initial policy is safe.

\begin{algorithm}[t]
\caption{Certified policy improvement}
\label{alg:main}
\begin{algorithmic}[1]
\State \textbf{Input:} safe initial policy $\theta_0$, $\delta$, $\eta$, step $\alpha$, episodes $E$
\For{$k=0,1,2,\dots$}
  \State Run $\pi_{\theta_k}$ for $E$ episodes; add the observed increments to the data
  \State Estimate $\nabla J(\theta_k)$ from the same episodes
  \State $\theta'\gets\theta_k-\alpha\,\widehat{\nabla J}(\theta_k)$
  \While{$B^{\mathrm u}_{\theta'}>\delta$} \Comment{at most $m$ times}
    \State $\theta'\gets(\theta'+\theta_k)/2$
  \EndWhile
  \State $\theta_{k+1}\gets\theta'$ if $B^{\mathrm u}_{\theta'}\leq\delta$, else $\theta_{k+1}\gets\theta_k$
\EndFor
\end{algorithmic}
\end{algorithm}

\begin{assumption}\label{as:seed}
The initial policy satisfies $p_{\theta_0}(x_0)\leq\delta$ and uses every action in every state with positive probability.
\end{assumption}

The second part ensures that every action is tried, so that its state-action CGF can be estimated.

\begin{excont}
The policy $\theta_0=0.05$ uses both modes in every state and has $p_{\theta_0}(x_0)=4.1\times10^{-11}$, so it satisfies Assumption~\ref{as:seed}.
\end{excont}

\begin{proposition}\label{prop:safe}
Under Assumptions~\ref{as:phi}--\ref{as:seed}, with probability at least $1-\eta$, every policy run by Algorithm~\ref{alg:main} satisfies $p_{\theta_k}(x_0)\leq\delta$.
\end{proposition}

\begin{proof}
Let $\Omega_\eta$ be the event of Lemma~\ref{lem:ucb}, so that $\Pp(\Omega_\eta)\geq1-\eta$. On $\Omega_\eta$, \eqref{eq:ucb} holds for every sample size, so whatever data the algorithm has collected, $\kappa^{\mathrm u}\geq\kappa$ for every cell, action and $\lambda\in\Lambda_{\mathrm g}$. Then $\kappa^{\mathrm u}_\theta\geq\kappa_\theta$, and Corollary~\ref{cor:cramer} or~\ref{cor:cells} gives $p_\theta(x_0)\leq B^{\mathrm u}_\theta$ for every $\theta$ at once. We show by induction on $k$ that $p_{\theta_k}(x_0)\leq\delta$ on $\Omega_\eta$. For $k=0$ this is Assumption~\ref{as:seed}. The next policy $\theta_{k+1}$ is either $\theta_k$, which satisfies the constraint by the induction hypothesis, or a $\theta'$ with $B^{\mathrm u}_{\theta'}\leq\delta$, computed from the data collected up to round $k$, before $\pi_{\theta'}$ is run. On $\Omega_\eta$ the second case gives $p_{\theta'}(x_0)\leq B^{\mathrm u}_{\theta'}\leq\delta$. The event $\Omega_\eta$ is the same for all rounds, so no union bound over the rounds is needed, although the data and the tested policies depend on the earlier rounds.
\end{proof}

The cost $J$ is not a rare quantity, so its gradient can be estimated in the usual way from the episodes. In Example~\ref{ex:chain} we use the score-function estimator with the cost-to-go of each step and its mean as baseline. Algorithm~\ref{alg:main} needs no multiplier and no gradient of the certificate. The iterates stay in the certified set, which grows as data are added.

\begin{remark}[Gradient of the certificate]
If a gradient of the certificate is needed, for example for a barrier method, it follows from \eqref{eq:polcgf}, $\nabla_\theta\kappa_\theta(x;\lambda)=\E_{u\sim\tilde\pi^\lambda_\theta(\cdot\mid x)}[\nabla_\theta\log\pi_\theta(u\mid x)]$ with the tilted policy $\tilde\pi^\lambda_\theta(u\mid x)=\pi_\theta(u\mid x)e^{\kappa(x,u;\lambda)-\kappa_\theta(x;\lambda)}$. It is a one-step expectation and needs no failures.
\end{remark}

\begin{remark}[Why the constraint is for $x_0$]
The certificate bounds only the failure probability from $x_0$, as such constraints do not satisfy Bellman's principle. In the example of \cite{haviv1996constrained}, the risk taken on a branch that is not followed limits what is allowed on the other branch, so the best action at an intermediate state depends on where the system started. At finite horizon this happens in general, because the failure probability always depends on the initial state. Algorithm~\ref{alg:main} therefore checks whole policies from $x_0$ and never checks single decisions from intermediate states with the same $\delta$.
\end{remark}

\section{Numerical Results}\label{sec:num}
We return to Example~\ref{ex:chain}. All numbers below are produced by the accompanying notebook, available at \url{https://github.com/rahul-misra/Rare-Event-Constrained-Safe-Reinforcement-Learning-with-Large-Deviations-Principle}. The true model is used only to check the results.

\begin{figure*}
\centering
\includegraphics[width=0.85\textwidth]{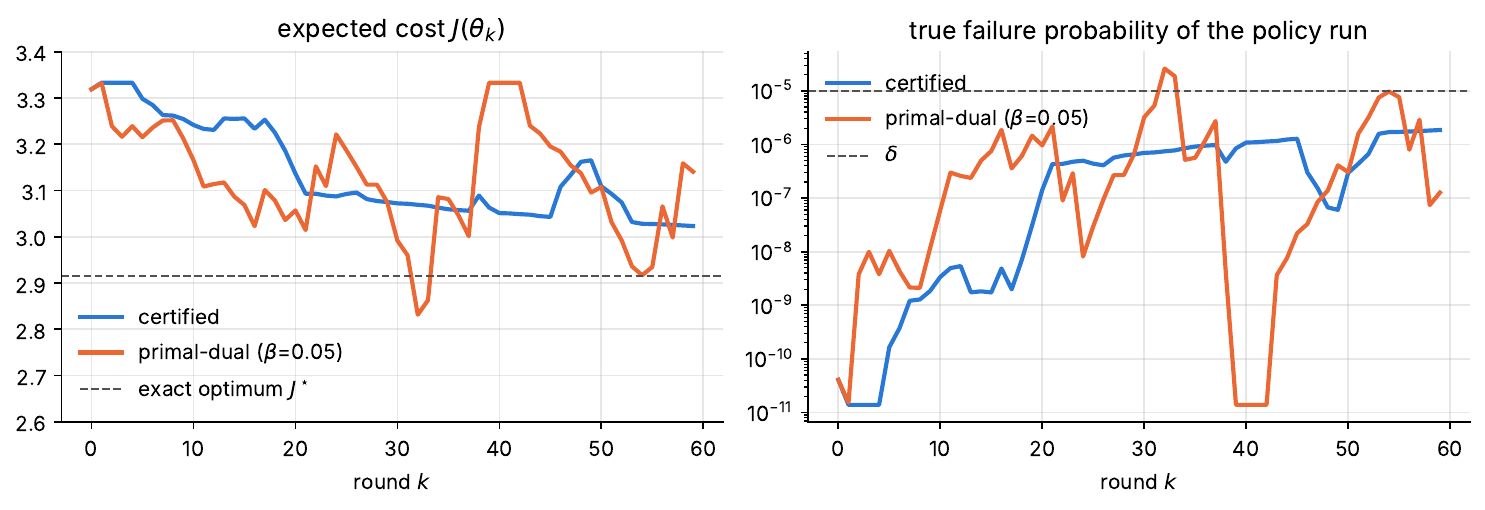}
\caption{Learning on Example~\ref{ex:chain}. Left: expected cost $J(\theta_k)$ of the policy run in round $k$. Right: true failure probability of the policy run in round $k$ (computed from the true model, for checking only). Algorithm~\ref{alg:main} (blue) moves only when the data allows it and stays below $\delta$. The primal--dual method (orange, $\beta=0.05$) moves first and corrects later.}
\label{fig:learning}
\end{figure*}

\noindent\textbf{Validity and tightness.} The certificate \eqref{eq:cramerbound} upper-bounds $p_\theta(x_0)$ for all $25$ combinations of $\theta\in\{0,0.25,0.5,0.75,1\}$ and $T\in\{20,30,60,100,1000\}$, with ratios from $1.09$ to $3.46$ (Table~\ref{tab:cert}). For $\theta=1$ with $x_0=0.1N$ and $T=6N$, $-\tfrac1N\log p$ equals $0.3993$, $0.3659$ and $0.3649$ at $N=20$, $100$ and $400$, while $-\tfrac1N\log B_\theta=0.3649$ for every $N$.

\begin{table}[t]
\centering
\caption{Exact failure probability and certificate \eqref{eq:cramerbound}, $N=20$, $x_0=2$.}
\label{tab:cert}
\small
\begin{tabular}{ccccc}
\toprule
$\theta$ & $T$ & $p_\theta(x_0)$ & $B_\theta$ & $B_\theta/p_\theta$\\
\midrule
$0$    & $30$  & $8.76\times10^{-12}$ & $1.46\times10^{-11}$ & $1.66$\\
$0$    & $100$ & $1.36\times10^{-11}$ & $1.49\times10^{-11}$ & $1.09$\\
$0.75$ & $30$  & $8.88\times10^{-7}$  & $2.85\times10^{-6}$  & $3.21$\\
$0.75$ & $100$ & $9.92\times10^{-6}$  & $1.47\times10^{-5}$  & $1.48$\\
$1$    & $30$  & $1.26\times10^{-5}$  & $4.35\times10^{-5}$  & $3.46$\\
$1$    & $100$ & $3.08\times10^{-4}$  & $6.82\times10^{-4}$  & $2.21$\\
\bottomrule
\end{tabular}
\end{table}

\noindent\textbf{Certificate from data.} Table~\ref{tab:data} shows the certificate with CGFs estimated by Lemma~\ref{lem:ucb} ($\eta=0.05$, a grid of $601$ points on $[0,3]$) from $n$ steps of the policy $\theta=0.5$, over $200$ independent data sets. No failure was observed in any data set. If no policy is certified, we keep $\theta=0$. The certified policy grows with $n$ towards the policy allowed by the exact CGFs, and the policy that was kept satisfied the constraint in every data set. Fig.~\ref{fig:bound} shows that the bound from one data set stays above $p_\theta(x_0)$ for all $\theta$.

\begin{table}[t]
\centering
\caption{Certificate from $n$ steps of the policy $\theta=0.5$ over $200$ data sets: share of data sets in which some policy is certified, median certified $\theta$ ($\theta=0$ if none), and median cost above the optimum $J=2.914$.}
\label{tab:data}
\small
\begin{tabular}{cccc}
\toprule
$n$ & certified & median $\theta$ & cost above optimum\\
\midrule
$10^3$ & $42\%$  & $0$     & $14.4\%$\\
$10^4$ & $100\%$ & $0.466$ & $7.9\%$\\
$10^5$ & $100\%$ & $0.638$ & $3.8\%$\\
exact CGF & -- & $0.727$ & $0.9\%$\\
\bottomrule
\end{tabular}
\end{table}

\noindent\textbf{Learning.} We run Algorithm~\ref{alg:main} for $60$ rounds of $E=400$ episodes with $\alpha=0.05$, starting from $\theta_0=0.05$, for which $p_{\theta_0}(x_0)=4.1\times10^{-11}$. For comparison we run a primal--dual method on the same certificate and the same gradient estimate, with multiplier step $\beta\in\{0.05,0.5\}$ (Fig.~\ref{fig:learning}). Algorithm~\ref{alg:main} ends $3.7\%$ above the exact optimum, and every policy it ran was checked before it was run. The primal--dual method ran $20$ and $11$ policies without a valid certificate at the time, and in both cases two of them had a true failure probability above $\delta$, up to $2.6\times10^{-5}$.

\noindent\textbf{State-dependent damage.} We change Example~\ref{ex:chain} so that the damage accelerates near failure: $q=0.2$ for $x<12$ and $q=0.45$ for $x\geq12$, so that $p=5.13\times10^{-8}$ without a horizon. With cells $A=\{x<12\}$ and $B=\{x\geq12\}$ and the exact state-dependent Lundberg coefficients, Corollary~\ref{cor:cells} gives $1.92\times10^{-7}$, $3.7$ times the true value, while Corollary~\ref{cor:cramer} gives $2.7\times10^{-2}$. In $2\times10^5$ episodes of normal operation the damage never went above $10$, so cell $B$ has no data and gets $s_B=0$. The certificate from data is $1.24\times10^{-6}$, and it remains valid even if the up-probability in cell $B$ is $0.99$. A transition model fitted to the same data, with the unseen states given the average up-probability, predicts $1.4\times10^{-11}$, which is overly optimistic (hence unsafe) compared to the ground truth.

\section{Conclusion}
In contrast to the rare event simulation literature, the goal of this paper was not to estimate the probability of the rare event occurrence, but to upper bound it with a suitable constraint. In practice, the controller cannot make failures frequent in a useful way, but it can learn the state-action cumulant generating functions of the damage from regular transitions. Through the Lundberg coefficient and its state-dependent versions, these give an upper bound on the failure probability that holds at every horizon via the large deviation principle, and can be checked for any policy before it is run. This leads to a simple learning algorithm that runs only certified policies. The method needs bounded (light-tailed) increments. For heavy-tailed distributions, the CGF does not exist, and other tools are needed \cite{blanchet2025}. Choosing the cells and the damage potential from data, and the behavior of the cost of certification as $\delta\to0$, are future work. Theorem~\ref{thm:cert} holds for any $W$ that satisfies \eqref{eq:super}, so a parametrized $W$, for example a neural network as in \cite{mathiesen2023}, could replace the certificates of Corollaries~\ref{cor:cramer} and~\ref{cor:cells}. This would be needed with function approximation, and is also future work.

\end{document}